\documentclass[letterpaper]{article} 
\usepackage{TrueMoE}  
\usepackage{times}  
\usepackage{helvet}  
\usepackage{courier}  
\usepackage[hyphens]{url}  
\usepackage{graphicx} 
\usepackage{natbib}  
\usepackage{caption} 
\usepackage{algorithm}
\usepackage{algorithmic}

\usepackage{newfloat}
\usepackage{listings}
\DeclareCaptionStyle{ruled}{labelfont=normalfont,labelsep=colon,strut=off} 
\floatstyle{ruled}
\newfloat{listing}{tb}{lst}{}
\floatname{listing}{Listing}
\usepackage{multirow}
\usepackage{booktabs}
\usepackage{utfsym}
\usepackage{makecell}
\usepackage{caption}
\usepackage{subcaption}
\usepackage{amsmath}
\usepackage{tikz,pgfplots,pgfplotstable}
\usepackage{color}
\usepackage{colortbl,hhline}
\usepgfplotslibrary[groupplots]

\title{TrueMoE: Dual-Routing Mixture of Discriminative Experts for \\Synthetic Image Detection}
\author{
    Laixin Zhang$^{1}$\equalcontrib, Shuaibo Li$^{2}$\equalcontrib, Wei Ma$^{1}$\thanks{Corresponding author.}, Hongbin Zha$^{3}$
}
\affiliations{
   \small $^{1}$Beijing University of Technology\\
    \small $^{2}$The Hong Kong University of Science and Technology (Guangzhou)\\
    \small $^{3}$Peking University\\
    {\small E-mail: \texttt{mawei@bjut.edu.cn}}
}

\nocopyright
\usepackage{bibentry}

\begin{document}

\maketitle

\begin{abstract}
The rapid progress of generative models has made synthetic image detection an increasingly critical task. Most existing approaches attempt to construct a single, universal discriminative space to separate real from fake content. However, such unified spaces tend to be complex and brittle, often struggling to generalize to unseen generative patterns. In this work, we propose \textbf{TrueMoE}, a novel dual-routing Mixture-of-Discriminative-Experts framework that reformulates the detection task as a collaborative inference across multiple specialized and lightweight discriminative subspaces. At the core of TrueMoE is a Discriminative Expert Array (DEA) organized along complementary axes of manifold structure and perceptual granularity, enabling diverse forgery cues to be captured across subspaces. A dual-routing mechanism, comprising a granularity-aware sparse router and a manifold-aware dense router, adaptively assigns input images to the most relevant experts. Extensive experiments across a wide spectrum of generative models demonstrate that TrueMoE achieves superior generalization and robustness.
\end{abstract}


\section{Introduction}

\begin{figure}[!t]
\centering
  \includegraphics[width=0.73\linewidth]{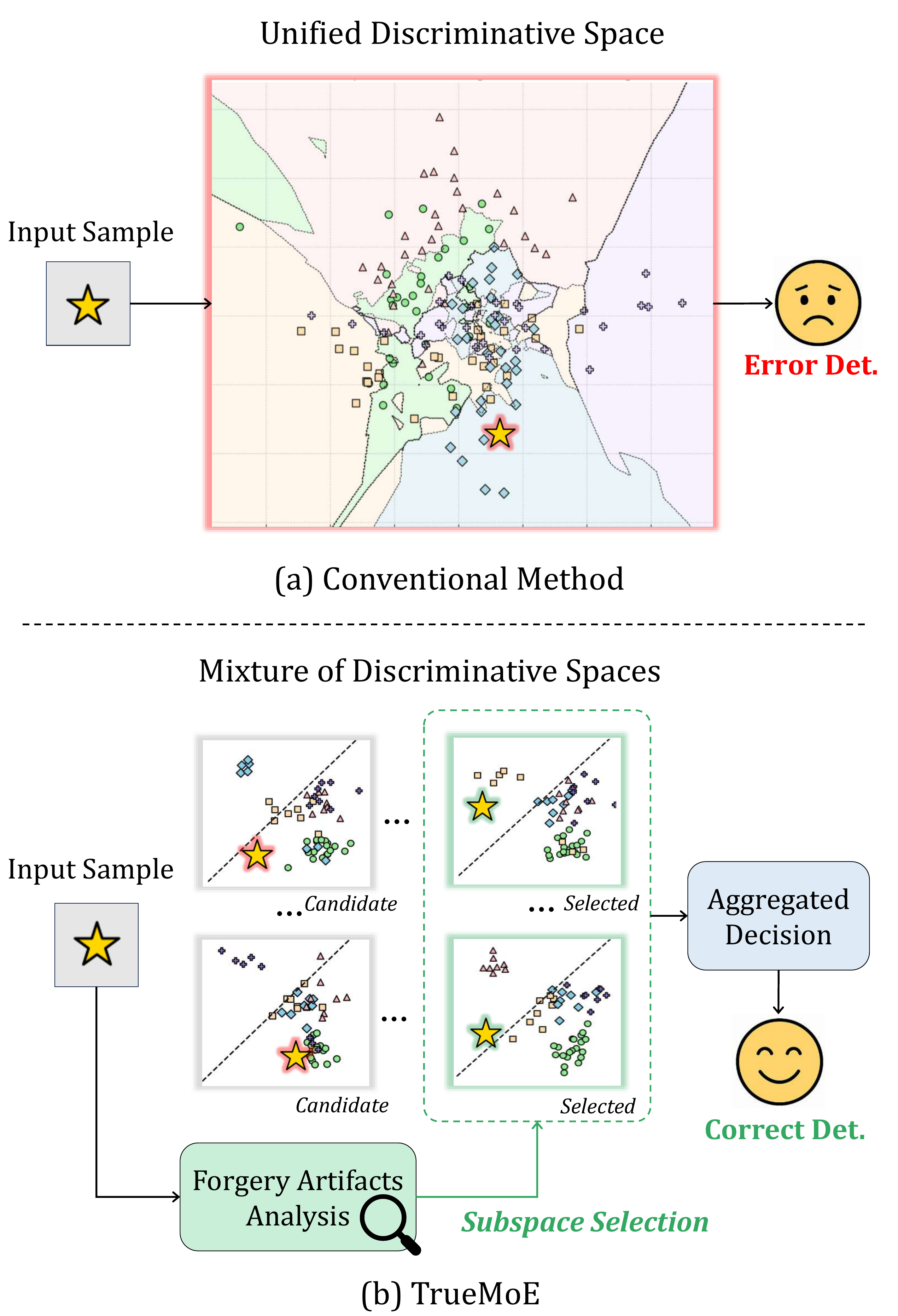}
  \vspace{-2mm}
  \caption{Conceptual comparison of detection paradigms.
(a) Conventional methods map all samples into a unified discriminative space without considering generative origin, often leading to misclassification.
(b) TrueMoE performs artifact-guided analysis to route inputs into specialized subspaces, improving generalization and classification accuracy. Red and green stars indicate misclassified and correctly classified input samples, respectively.}
  \vspace{-2mm}
  \label{fig:1}
  
\end{figure}

Recent advances in generative models, such as GANs~\cite{intro_301,intro_302,intro_303,intro_304} and diffusion models~\cite{intro_307,intro_308,intro_309,intro_312}, have enabled the synthesis of highly realistic and diverse visual content. These models benefit creative industries but also introduce serious security and ethical risks. Synthetic images can be exploited to disseminate misinformation~\citep{xu2023combating}, falsify evidence~\citep{vincent2023pope}, and manipulate public opinion~\cite{kietzmann2020deepfakes}. As their realism improves, traditional detection signals become increasingly unreliable, especially when facing unseen generators or post-processing distortions. This underscores the need for detection systems that can generalize across diverse generative patterns and maintain robustness in open-world scenarios.

Prior work in synthetic image detection has predominantly evolved along two lines. Early approaches adopt supervised learning to extract discriminative features from spatial~\cite{method_14, tan2024rethinking, drct} or frequency domains~\cite{method_114, method_26}, often tailored to specific generative models. More recently, methods have shifted toward leveraging large-scale pre-trained representations~\cite{method_12,fatformer,method_11}, such as CLIP~\cite{radford2021learning} embeddings or features from diffusion models, combined with lightweight classifiers to enhance generalization.  While these approaches have demonstrated promising results, they share a common design principle: they attempt to construct a unified feature space that can separate all real and fake samples via a global decision boundary. This paradigm simplifies the detection process but implicitly assumes that forgery artifacts, regardless of their origin, are separable in a single space. 

Building upon the unified-space paradigm, many existing detection approaches assume that all samples can be processed through a universal decision space, without accounting for the generative origin or structural variability of synthetic artifacts. In reality, different generators produce markedly different artifact distributions, both in terms of latent manifold structures and perceptual appearances. Without explicit analysis of these differences, a single global feature space may lack flexibility to capture the diversity of forgery artifacts, potentially affecting generalization in more open or challenging scenarios, as shown in Figure \ref{fig:1}(a).

In this work, we propose a fundamentally different formulation. Instead of treating detection as a global binary classification task within a single space, we decompose it into multiple subspaces, each handled by an expert. This leads to TrueMoE, a novel Mixture-of-Discriminative-Experts framework designed for adaptive, generalizable synthetic image detection. At the core of TrueMoE lies the Discriminative Expert Array (DEA), a structured set of experts, each trained within a specialized subspace to capture complementary forgery cues. This design emphasizes analysis before inference: instead of applying a fixed feature extractor to all inputs, we first assess the intrinsic characteristics of each image to guide expert selection. This paradigm shift enables more robust modeling of cross-model artifacts and provides greater interpretability, as illustrated in Figure\ref{fig:1} (b).

DEA is organized along two functionally independent dimensions: manifold structure and perceptual granularity. The manifold axis reflects the observation that artifacts generated by different models tend to reside on distinct manifolds, influenced by architectural design, training objectives, and latent space properties~\cite{wang2024trace, method_26}. In contrast, the granularity axis captures the intuition that, although forgery cues may theoretically appear across multiple perceptual levels, each image typically presents a dominant level of abstraction at which these cues are most effectively detectable~\cite{aeroblade, guo2023hierarchical}. Prior studies have shown that the detectability of synthetic traces is often concentrated within a particular representation layer depending on the semantic content, local image structure and generative artifacts~\cite{wang2022m2tr, li2024improving}. Thus, we adopt a sparse granularity strategy: instead of fusing responses across all levels, which may dilute salient signals or introduce redundancy, we activate only the experts corresponding to the most informative perceptual level. This reinforces the notion that the most discriminative features tend to emerge at specific representation depths unique to each image. 

To implement this dual-axis assignment, we introduce two complementary routing modules. The Granularity Routing Module (GRM) performs sparse selection by estimating the optimal perceptual scale per input and activating a single corresponding expert. Meanwhile, the Hybrid Manifold Routing Module (HMRM) extracts a manifold-aware latent representation and performs dense routing across manifold-specific experts. Together, these modules enable adaptive, analysis-guided expert activation and facilitate robust detection across diverse generative domains.

Our contribution can be summarized as follows:
\begin{itemize}
\item We reformulate synthetic image detection as a multi-view subspace classification task, moving beyond the conventional unified-space assumption. Our method models forgery cues via a structured ensemble of specialized experts.
\item We propose TrueMoE, a Mixture-of-Discriminative-Experts framework that models forgery artifacts through the Discriminative Expert Array (DEA), a two-dimensional ensemble organized along manifold structure and perceptual granularity.
\item We design a dual-axis routing strategy, including a Granularity Routing Module (GRM) for sparse selection of the most informative perceptual level, and a Hybrid Manifold Routing Module (HMRM) for dense expert aggregation across manifold-aware subspaces.
\item Extensive experiments across diverse generative models demonstrate that TrueMoE consistently outperforms state-of-the-art methods in both detection accuracy and generalization robustness.
\end{itemize}

\begin{figure*}[h]
  \centering
  \includegraphics[width=0.85\linewidth]{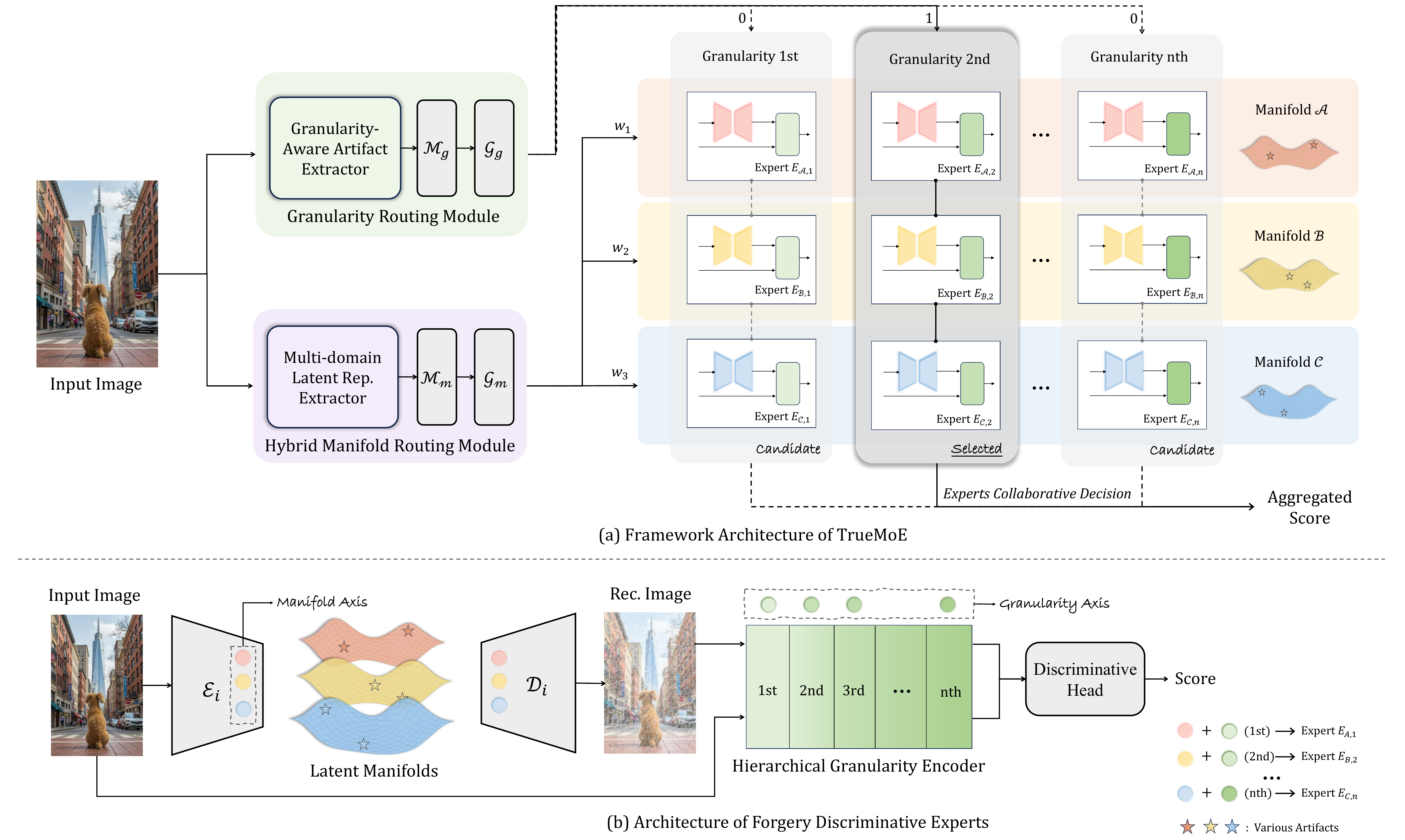}
 \vspace{-1mm}
  \caption{
  Overview of TrueMoE. (a) Dual routing mechanism enables adaptive expert selection based on both artifact granularity and manifold structure. (b) Architecture of a single expert, designed to model diverse manifolds and perceptual granularities.}
  \label{fig:2}
\end{figure*}

\section{Related work}

\subsection{Synthetic Image Detection}
The rapid progress of generative models has spurred the emergence of various synthetic image detection methods, which can be broadly classified into two categories: feature embedding-based and reconstruction-based approaches.  
Feature embedding-based methods aim to extract discriminative representations by designing task-specific networks or leveraging pre-trained models. For example, CNNSpot~\cite{method_14} trains a simple ResNet solely on augmented ProGAN images~\cite{intro_301} and generalizes well across various GANs. UniFD~\cite{method_12} employs CLIP features and trains a linear classifier, while CO-SPY~\cite{co-spy} also adopts CLIP to model high-level semantics. FreDect~\cite{method_26} differs by incorporating the discrete cosine transform (DCT) to capture frequency-domain artifacts.
Reconstruction-based methods exploit the insight that synthetic images align more closely with the latent space of generative models, making them easier to reconstruct. DIRE~\cite{method_11}, for instance, uses a pre-trained ADM~\cite{intro_307} for reconstruction, using the reconstruction error for detection.  More recently, AEROBLADE~\cite{aeroblade} revitalizes this paradigm by adopting LDM-based autoencoders and computing LPIPS~\cite{lpips} without requiring costly diffusion processes.  Nonetheless, these methods operate within a fixed detection space, which limits their generalization to unseen generators. In contrast, our proposed TrueMoE dynamically selects experts based on input characteristics, leading to improved generalization performance.

\subsection{Knowledge Fusion}
Knowledge fusion from diverse sources is a classical strategy for building robust AI models.  
Traditional ensemble methods~\cite{zhou2021ensemble, zounemat2021ensemble} improve generalization and stability by training multiple models on different sub-datasets and aggregating their predictions.  
 Inspired by this, recent advancements in large language models~\cite{2.3.3, lepikhin2020gshard, liu2024deepseek} have demonstrated that merging multiple models can lead to strong performance across diverse tasks.  
 In addition, Mixture-of-Experts (MoE)~\cite{moe} offers a scalable and effective framework for knowledge integration. By selectively activating a subset of specialized experts based on the input, MoE enables the decomposition of complex problems into simpler subtasks while balancing model capacity and computational efficiency.  
It has shown remarkable success in natural language processing tasks such as translation~\cite{shazeer2017outrageously, costa2022no}, code generation~\cite{jiang2024mixtral, dai2024deepseekmoe}, and open-domain QA~\cite{du2022glam, artetxe2021efficient}. More recently, it has also achieved success in computer vision via sparse expert activation~\cite{2.3.2, pmoe}.
Building on these insights, we extend the knowledge fusion paradigm to the domain of synthetic image detection. Unlike prior methods that rely on a unified detection space, we aim to construct a more expressive forgery detection space by jointly modeling manifold diversity and perceptual granularity.

\section{Method}
In this section, we introduce TrueMoE, a Mixture-of-Experts framework for synthetic image detection.
An overview of its architecture is illustrated in Figure \ref{fig:2}.

\begin{figure*}[h]
  \centering
  \includegraphics[width=0.85\linewidth]{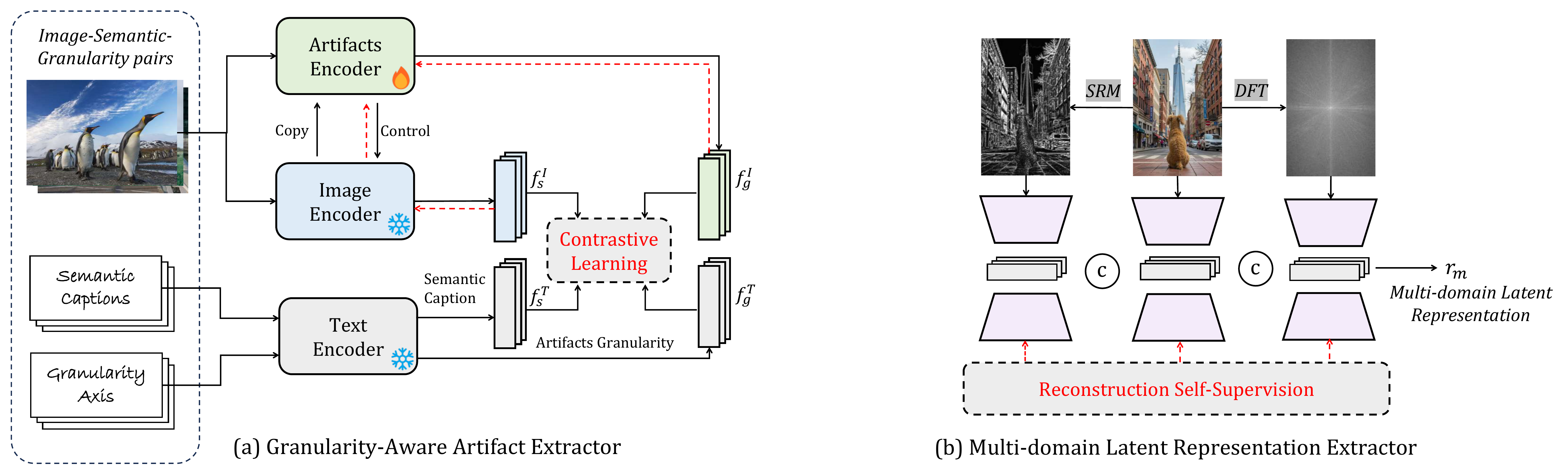}

  \caption{(a) The  Granularity-Aware Artifact Extractor is trained with contrastive supervision using image-text-granularity triplets, separating semantic and artifact embeddings.
(b) The Multi-domain Latent Representation Extractor is pretrained via reconstruction from RGB, SRM, and DFT domains to capture diverse latent representation.}
\label{fig:3}
\end{figure*}
\subsection{Preliminaries}
Mixture-of-Experts (MoE) is a deep learning paradigm that integrates multiple expert networks through dynamic routing. Its core idea is to adaptively select relevant experts through a gating mechanism. A typical MoE model comprises a set of experts $\{E\}$ (e.g., lightweight subnetworks or linear layers) and a gating network $\mathcal{G}$ that assigns selection weights. The gating network typically includes a linear layer, ReLU activation, and a Softmax function. In dense MoE, all experts are activated, and their outputs are aggregated based on the gating weights. Given an input $x$, the output is computed as:  

\begin{equation}
\mathcal{G}(h)=\operatorname{softmax}(g(h))
\end{equation}
\begin{equation}
F \left(x, h\right)
= \sum_{i=0}^N \mathcal{G}(h)_i E_i\left(x\right)
\end{equation}
where $h$ denotes the feature representation of $x$, and $g(\cdot)$ denotes the gating logits computed by the gating network. Unlike dense MoE, sparse MoE activates only the top-$k$ experts for each input. This design improves scalability by increasing model capacity without proportional computational cost. The sparse gating function is defined as:

\begin{equation}
  \mathcal{G}(h)=\operatorname{softmax}(TopK(g(h)+R_{\text{noise}}, k))
\end{equation}

\begin{equation}
  TopK(g(h), k)_i=\left\{\begin{array}{l}
g(h)_i,  i \in topk(g(h)) \\
-\infty, \text { otherwise}
\end{array}\right.
\end{equation}
where $R_{\text{noise}}$ is a commonly used regularization technique to enhance the training stability of sparse gating networks.

\subsection{Discriminative Expert Array}
Existing methods often assume a unified feature space where all forgery artifacts are separable, but this oversimplification fails to capture the diversity of artifacts from different generators.
To address this, we propose the Discriminative Expert Array (DEA), a structured expert ensemble designed to enhance generalization and adaptability in synthetic image detection. 
 Inspired by AEROBLADE~\cite{aeroblade}, we decompose the detection space along two synergistic axes: \textbf{manifold structure}, which captures generator-specific latent representations, and \textbf{perceptual granularity}, which models inconsistencies across different abstraction levels.
As illustrated in Figure~\ref{fig:2}(a), DEA arranges experts in a two-dimensional grid, each expert specialized to a unique combination of manifold and granularity levels. Specialization is achieved by varying either the encoder-decoder architecture or the granularity encoder.

\noindent \textbf{Forgery Discriminative Expert.}
 Each forgery discriminative expert in DEA is a lightweight module consisting of three components: an encoder-decoder pair $(\mathcal{E}_i, \mathcal{D}_i)$ , a Hierarchical Granularity Encoder (HGE) $\mathcal{H}$, and a Discriminative Head $\mathcal{P}$, as illustrated in Figure~\ref{fig:2}(b). To reduce computation, only the discriminative head is trainable while all other components remain frozen.

From the manifold perspective, generative models tend to produce distinct latent structures.
To model this diversity, each expert is assigned a frozen encoder-decoder pair pretrained on a latent diffusion model.
Given an input $x$, the encoder maps it to a manifold-specific latent representation $z_i = \mathcal{E}_i(x)$, which the decoder reconstructs as $x'_i = \mathcal{D}_i(z_i)$.
This reconstruction reveals generator-specific discrepancies that serve as forensic cues.
Synthetic images, being closer to the learned manifolds, generally yield lower reconstruction errors than real ones—an asymmetry that supports forgery detection.
Compared to training individual models, the frozen autoencoder design further enables efficient and compact implementation across multiple experts.

From the granularity perspective, different images exhibit artifact traces of varying intensity across perceptual scales~\cite{aeroblade}.
 To capture such granularity-dependent traces, we implement the HGE using multiple layers from a pre-trained backbone (e.g., VGG-16). Shallow layers are sensitive to low-level textures, while deeper layers capture higher-level inconsistencies. For the expert $E_{i,j}$, given the original image $x$ and its reconstruction $x'_i$, we compute the Granularity-Aware Discrimination Feature (GDF) at the $j$-th level of the HGE module as:  
\begin{equation}
\text{GDF}_j(x) = \text{Agg}\left(\mathcal{H}(x)_j, \mathcal{H}(x'_i)_j\right)
\end{equation}
where $\text{Agg}(\cdot)$ computes the element-wise residual between feature maps.
This residual highlights subtle inconsistencies that are indicative of forgery.
The GDF is then fed into a lightweight head to generate the final prediction.

\subsection{Dual Routing Mechanism}
\label{3.3}
TrueMoE adopts a dual-routing strategy that assigns inputs to experts based on estimated manifold structures and perceptual granularities.
A Hybrid Manifold Routing Module and a Granularity Routing Module independently estimate the latent structure and perceptual scale, facilitating precise expert selection for robust synthetic image detection.

\noindent \textbf{Granularity Routing Module.} 
The Granularity Routing Module (GRM) assigns each input to the most suitable expert group by estimating its artifact granularity, which reflects the abstraction level where forgery traces are most perceptually salient.
To extract such cues, we introduce the Granularity-Aware Artifact Extractor (GAE), a lightweight encoder trained to isolate scale-specific forgery cues while suppressing high-level semantics.

As shown in Figure~\ref{fig:3}(a), we construct a CLIP-style contrastive pretraining framework to disentangle artifact-related granularity from semantic content.
Each image is paired with a semantic caption from BLIP~\cite{blip} and a granularity label indicating its optimal perceptual scale (e.g., “First-level granularity” to “Sixth-level granularity”).
Given an input image $I$, its semantic caption $T_s$, and granularity label $T_g$, the following embeddings are generated:

\begin{equation}
\begin{split}
f_s^I &= \mathcal{E}_{v}(I) + \mathrm{GAE}(I), \quad\;\; f_g^I = \mathrm{GAE}(I), \\
f_s^{T} &= \mathcal{E}_{t}(T_{s}), \quad\quad\quad\quad\quad\;\; f_g^{T} = \mathcal{E}_{t}(T_{g}).
\end{split}
\end{equation}
where $\mathcal{E}_v$ and $\mathcal{E}_t$ are frozen encoders, and $\mathrm{GAE}$ is trained with LoRA-based adaptation~\cite{lora} to reduce training overhead and mitigate overfitting. The total loss combines two contrastive objectives:

\begin{equation}
\begin{gathered}
\mathcal{L}_{total} = \mathcal{L}_{con}(f_s^I, f_s^T) + \mathcal{L}_{con}(f_g^I, f_g^T),
\\
\hspace*{-0.9mm}
\mathcal{L}_{\text{con}}(f^I, f^T) = -\frac{1}{N}\sum_{i=1}^{N} \log \frac{\exp((f_i^I)^\top f_i^T / \tau)}{\sum_{j=1}^{N} \exp((f_i^I)^\top f_j^T / \tau)}.
\end{gathered}
\end{equation}
where $N$ is the batch size and $\tau$ is the temperature parameter.

This training enforces clear separation between semantic and granularity representations, enabling GRM to route inputs based on artifact scale rather than content.
After pretraining, GAE is frozen and used at inference to compute granularity-aware embeddings.
A sparse top-$k$ routing strategy ($k = 1$ in practice) is applied, routing each image to its most relevant expert group $\{E_{i,j}\}_{j=t}$ according to the predicted granularity level.

\noindent \textbf{Hybrid Manifold Routing Module.}
Complementing the GRM, the Hybrid Manifold Routing Module (HMRM) estimates the manifold structure of each input to support granular expert activation. Due to architectural and training differences, images from different generative models lie on distinct manifolds. Modeling these differences enhances both routing precision and generalization.
To this end, we design the Multi-Domain Latent Representation Extractor (MLRE) to capture complementary features from the spatial domain (RGB), local noise residuals (SRM ~\cite{srm}) and frequency domain (DFT), as shown in Figure~\ref{fig:3}(b). Inspired by~\cite{method_114,li2024unionformer}, this multi-view design enables more comprehensive manifold estimation, guiding HMRM to activate the most relevant experts.

MLRE employs three parallel encoder-decoder branches, each trained using a self-supervised reconstruction task to learn domain-specific manifolds.
Formally, for input image $x$, each branch reconstructs its corresponding signal as:
\begin{equation}
\begin{gathered}
\tilde{x}_r = \mathcal{D}_r(\mathcal{E}_r(x)), \quad \tilde{x}_s = \mathcal{D}_s(\mathcal{E}_s(\mathrm{SRM}(x))), \\
\tilde{x}_f = \mathcal{D}_f(\mathcal{E}_f(\mathrm{DFT}(x))).
\end{gathered}
\end{equation}
where $\mathcal{E}_{\{r,s,f\}}$ and $\mathcal{D}_{\{r,s,f\}}$ denote the encoder and decoder for each domain.
The training objective minimizes the pixel-wise reconstruction loss across all domains:

\begin{equation}
\begin{gathered}
\mathcal{L}_{rec}^{(o)} = \frac{1}{N} \sum_{i=1}^{N} \|x - \tilde{x}_o\|^2, \quad o \in \{r, s, f\},
\end{gathered}
\end{equation}

After pretraining, the decoders are discarded and the encoder weights are frozen. The final multi-domain latent representation $r_m$ is obtained by concatenating the outputs of the three domain-specific encoders:
\begin{equation}
r_m = \text{Concat}(\mathcal{E}_r(x), \mathcal{E}_s(SRM(x)), \mathcal{E}_f(DFT(x)))
\end{equation}

This multi-domain latent representation is subsequently fed into the manifold routing gate, which generates routing scores to facilitate collaborative decision-making among the experts selected by the GRM. 
By integrating diverse domain-specific cues, this dense routing strategy improves generalization to unseen generative distributions and ensures robust detection under real-world post-processing such as compression and resizing.

\begin{table*}[t]
  \centering
  
  \resizebox{\textwidth}{!}{
    \begin{tabular}{lccccccccccccccccccc}
    \toprule
    \multirow{3}*{Methods}  & \multirow{3}*{Ref} & \multicolumn{8}{c}{GAN}   & \multicolumn{9}{c}{Diffusion}  & \multirow{3}*{mAP} \\

    \cmidrule(r){3-10} \cmidrule(r){11-19}
     ~ & ~ & \makecell[c]{\small Pro-\\\small GAN} & \makecell[c]{\small Style-\\\small GAN} & \makecell[c]{\small Big-\\\small GAN} & \makecell[c]{\small Cycle-\\\small GAN} & \makecell[c]{\small star-\\\small GAN} & \makecell[c]{\small Gau-\\\small GAN} & \makecell[c]{\small Style-\\\small GAN2} & \makecell[c]{\\\small WFIR} & \makecell[c]{\\\small ADM}   & \makecell[c]{\\\small Glide} & \makecell[c]{\small Mid-\\\small journey} & \makecell[c]{\small SD\\\small v1.4} & \makecell[c]{\small SD\\\small v1.5} & \small \makecell[c]{\\\small VQDM}  & \makecell[c]{\small Wu-\\\small kong} & \makecell[c]{\small DALL\\ \small E2} & \makecell[c]{\small SD\\ \small -XL}&  \\
     \midrule
    CNNSpot & \small CVPR 2020  & 100.0 & 99.54 & 84.51 & 93.47 & 98.15 & 89.49 & 99.06 & 83.10 & 71.07 & 66.16 & 55.91 & 56.88 & 57.25 & 61.93 & 52.85 & 50.54 & 71.17 & 75.95 \\
    FreDect & \small ICML 2020 & 99.99 & 88.98 & 93.62 & 84.78 & 99.49 & 82.84 & 82.54 & 44.46 & 63.71 & 54.72 & 47.26 & 38.51 & 38.42 & 86.01 & 40.44 & 38.20 & 49.43 & 66.67 \\
    GramNet & \small CVPR 2020 & 100.0 & 94.49 & 62.34 & 74.81 & 100.0 & 55.28 & 99.37 & 51.15 & 56.98 & 55.08 & 58.70 & 63.02 & 63.18 & 54.05 & 60.85 & 53.74 & 58.16 & 68.31 \\
    LGrad & \small  CVPR 2023 & 100.0 & 97.91 & 89.11 & 93.78 & 99.98 & 91.32 & 98.21 & 51.11 & 66.96 & 82.41 & 73.50 & 63.26 & 63.69 & 71.77 & 61.20 & 83.92 & 73.28 & 80.08 \\
    UniFD & \small CVPR 2023 & 100.0 & 97.48 & 99.27 & 99.80 & 99.37 & 99.98 & 97.71 & 94.22 & 89.80 & 88.04 & 49.72 & 68.63 & 68.07 & 97.53 & 78.44 & 66.06 & 67.59 & 85.98 \\
    DIRE  & \small ICCV 2023 & 99.92 & 98.92 & 75.68 & 81.87 & 99.65 & 72.07 & 98.96 & 58.59 & 73.32 & 69.65 & 69.47 & 65.36 & 65.02 & 62.07 & 64.03 & 58.20 & 53.45 & 74.48 \\
    AEROBLADE  & \small  CVPR 2024 & 46.48 & 40.18 & 42.14 & 40.87 & 43.38 & 40.71 & 37.38 & 31.20 & 87.42 & 97.96 & 99.84 & 98.68 & 98.87 & 78.42 & 99.07 & 98.69 & 98.74 & 69.41 \\

    NPR & \small CVPR 2024 & 99.95 & 99.86 & 84.40 & 97.83 & 100.0 & 81.73 & 99.99 & 67.62 & 79.14 & 86.55 & 83.84 & 84.37 & 84.38 & 80.84 & 77.63 & 79.56 & 87.72 & 86.79 \\
    FatFormer & \small CVPR 2024 & 100.0 & 99.75 & 99.98 & 100.0 & 100.0 & 100.0 & 99.92 & 98.48 & 91.73 & 95.99 & 62.76 & 81.12 & 81.09 & 96.99 & 85.86 & 81.84 & 86.95 & 91.91 \\
    DRCT  & \small ICML 2024 & 91.03 & 79.44 & 93.51 & 98.68 & 96.29 & 86.62 & 73.80 & 91.04 & 88.96 & 94.64 & 97.03 & 99.65 & 99.49 & 96.54 & 99.37 & 97.67 & 95.44 & 92.89 \\

    D$^3$& \small CVPR 2025 & 100.0   & 97.59 & 98.18 & 99.91 & 98.84 & 99.04 & 96.52 & 95.78 & 95.86 & 92.42 & 80.15 & 84.83 & 84.96 & 93.38 & 85.44 & 81.34 & 81.75 & 92.12 \\
    CO-SPY & \small CVPR2025 & 100.0   & 98.35 & 97.11 & 99.74 & 98.73 & 99.21 & 98.85 & 93.98 & 90.50  & 96.39 & 88.42 & 92.86 & 92.87 & 96.28 & 93.72 & 93.54 & 92.22 & \underline{95.46} \\
    \rowcolor{gray!20}Ours   & -     & 99.99 & 98.85 & 98.80  & 97.15 & 99.58 & 97.49 & 96.77 & 97.20 & 97.40  & 98.02 & 93.28 & 98.36 & 98.40  & 97.26 & 98.17 & 94.03 & 96.72 & \textbf{97.50} \\
    \bottomrule
    \end{tabular}%
    }
    \caption{Average precision (\%) comparisons with SOTA methods. Ours performance is highlighted in gray. The first and second rankings are shown in bold and underlined respectively.}
  \label{tab:1}%
\end{table*}%

\begin{table*}[t]
  \centering
  
  \resizebox{\textwidth}{!}{
    \begin{tabular}{lccccccccccccccccccc}
    \toprule
    \multirow{3}*{Methods}  & \multirow{3}*{Ref} & \multicolumn{8}{c}{GAN}   & \multicolumn{9}{c}{Diffusion}  & \multirow{3}*{mAcc} \\

    \cmidrule(r){3-10} \cmidrule(r){11-19}
     ~ & ~ & \makecell[c]{\small Pro-\\\small GAN} & \makecell[c]{\small Style-\\\small GAN} & \makecell[c]{\small Big-\\\small GAN} & \makecell[c]{\small Cycle-\\\small GAN} & \makecell[c]{\small star-\\\small GAN} & \makecell[c]{\small Gau-\\\small GAN} & \makecell[c]{\small Style-\\\small GAN2} & \makecell[c]{\\\small WFIR} & \makecell[c]{\\\small ADM}   & \makecell[c]{\\\small Glide} & \makecell[c]{\small Mid-\\\small journey} & \makecell[c]{\small SD\\\small v1.4} & \makecell[c]{\small SD\\\small v1.5} & \small \makecell[c]{\\\small VQDM}  & \makecell[c]{\small Wu-\\\small kong} & \makecell[c]{\small DALL\\ \small E2} & \makecell[c]{\small SD\\ \small -XL}&  \\
     \midrule
    CNNSpot & \small CVPR2020 &99.99 & 85.73 & 70.18 & 85.20 & 91.75 & 78.91 & 83.39 & 75.65 & 58.78 & 55.01 & 52.60 & 51.57 & 51.78 & 53.68 & 50.24 & 49.85 & 58.53 & 67.81 \\
    FreDect & \small ICML2020 & 99.36 & 78.02 & 81.98 & 78.77 & 94.62 & 80.56 & 66.19 & 46.45 & 64.68 & 55.44 & 46.89 & 40.04 & 40.38 & 78.96 & 41.54 & 34.65 & 51.28 & 63.52 \\
    GramNet & \small CVPR 2020 & 99.99 & 83.59 & 67.53 & 73.69 & 100.0 & 57.77 & 85.86 & 51.40 & 57.76 & 57.82 & 60.83 & 65.28 & 65.71 & 56.88 & 63.41 & 53.80 & 59.25 & 68.27 \\
    LGrad & \small CVPR 2023 & 99.76 & 89.61 & 82.03 & 85.54 & 98.17 & 80.28 & 85.97 & 51.25 & 61.69 & 70.47 & 67.51 & 63.09 & 63.39 & 68.75 & 58.26 & 68.50 & 67.00 & 74.19 \\
    UniFD & \small CVPR 2023 & 99.81 & 80.40 & 95.08 & 98.33 & 95.75 & 99.47 & 70.76 & 72.70 & 67.46 & 63.09 & 49.87 & 51.70 & 51.59 & 86.01 & 55.14 & 50.80 & 50.73 & 72.86 \\
    DIRE  & \small ICCV 2023 & 98.58 & 93.57 & 72.73 & 72.79 & 96.22 & 68.45 & 94.10 & 56.45 & 68.68 & 63.69 & 61.95 & 61.44 & 61.24 & 59.51 & 58.49 & 56.35 & 54.26 & 70.50 \\
    AEROBLADE  & \small CVPR 2024 & 46.96 & 39.92 & 42.33 & 41.79 & 52.68 & 42.56 & 35.62 & 20.50 & 53.14 & 51.29 & 51.73 & 53.63 & 53.08 & 53.32 & 53.71 & 87.05 & 71.10 & 50.02 \\

    NPR  & \small CVPR 2024 & 99.84 & 97.53 & 83.20 & 94.10 & 99.70 & 79.97 & 99.40 & 65.50 & 74.73 & 81.48 & 78.82 & 80.09 & 80.64 & 79.02 & 73.80 & 69.35 & 81.28 & 83.44 \\
    
    FatFormer & \small CVPR 2024 & 99.89 & 97.13 & 99.50 & 99.36 & 99.75 & 99.43 & 98.80 & 88.10 & 78.44 & 88.03 & 56.09 & 67.83 & 68.06 & 86.88 & 85.86 & 69.70 & 73.50 & 85.67 \\
    DRCT  & \small ICML 2024 & 76.83 & 71.58 & 82.68 & 94.51 & 63.13 & 78.39 & 67.75 & 63.10 & 79.42 & 89.18 & 91.50 & 95.01 & 94.41 & 90.03 & 94.68 & 92.55 & 89.90 & 83.21 \\

    D$^3$& \small CVPR 2025 & 99.89 & 90.84 & 97.84 & 98.35 & 97.60 & 99.79 & 88.91 & 84.86 & 77.36 & 74.21 & 70.44 & 73.73 & 73.82 & 89.72 & 72.32 & 70.07 & 71.13 & 84.17 \\
CO-SPY & \small CVPR2025 &99.96 & 97.83 & 90.57 & 99.14 & 94.06 & 93.29 & 98.62 & 70.26 & 75.17 & 86.44 & 87.23 & 85.77 & 85.73 & 81.02 & 80.95 & 75.84 & 89.01 & \underline{87.70} \\
    \rowcolor{gray!20}Ours   & -     & 99.99 & 94.94 & 93.76 & 91.30  & 90.28 & 87.42 & 86.40  & 88.12 & 87.09 & 90.17 & 86.06 & 92.55 & 92.57 & 89.51 & 90.64 & 89.40  & 91.34 & \textbf{90.68} \\
    \bottomrule
    \end{tabular}%
 
  }
  \caption{Accuracy (\%) comparisons with SOTA methods. The notations are consistent with Table \ref{tab:1}.}
   \label{tab:2}%
\end{table*}%

\section{Training Strategy}
To ensure efficient yet effective training, we adopt a streamlined training strategy that leverages pre-trained modules while minimizing redundant optimization.
We start by extracting granularity-aware artifact embeddings $\{f_g^I\}$ from all training images using the frozen GAE module. We then apply K-means clustering~\cite{kmean} to construct a discrete perceptual space, yielding six cluster centers $\{u_n\}$ and corresponding training subsets $\{D_t\}_{t=1}^{6}$. All experts ${E_{i,j}}$ are initially pre-trained jointly on the full dataset to acquire general discriminative capabilities, optimized using the standard binary cross-entropy loss $\mathcal{L}_d$.

Subsequently, each expert subset $\{E_{i,j}\}_{j=t}$ is fine-tuned on its corresponding cluster $D_t$, allowing it to specialize in a specific perceptual level.
As all components except the discriminative head are frozen during training,  this phase is lightweight and computation-friendly.

The dual-routing modules are trained on the full dataset.
The granularity-aware artifact embeddings $f_g^I$ and the multi-domain latent representation $r_m$ are fed into two independent mapping networks, $\mathcal{M}_g$ and $\mathcal{M}_m$, producing granularity and manifold features $f_g$ and $f_m$, respectively.
These are then routed through the corresponding networks $\mathcal{G}_g$ and $\mathcal{G}_m$.

Inspired by uncertainty learning~\cite{uncertain}, we define the routing loss as:
\begin{equation}
\mathcal{L}_{\text{router}} = \frac{1}{2f_m}(u_i - f_g)^2 + \frac{1}{2} \ln f_m
\end{equation}  
where $u_i$ denotes the assigned granularity center for the input.
This uncertainty-aware formulation encourages stable and accurate expert selection while mitigating overconfidence.
To further reduce routing bias and encourage fair expert participation, we introduce a load-balancing regularizer:
\begin{equation}
\mathcal{L}_{balance} = n \sum_{i=1}^{n} d_i \cdot p_i.
\end{equation}
where $d_i$ and $p_i$ denote the actual and predicted expert allocation distributions, respectively.
The final objective integrates detection, routing, and regularization:
\begin{equation}
\mathcal{L} = \mathcal{L}_d + \alpha \cdot \mathcal{L}_{router} + \beta \cdot \mathcal{L}_{balance}.
\end{equation}

\section{Experiments}
\subsection{Experimental Setup}
\noindent \textbf{Dataset.}
We follow the standard protocol as adopted in prior works~\cite{method_14, method_12}.
Specifically, TrueMoE is trained exclusively on synthetic images generated by ProGAN. The evaluation is conducted on a comprehensive cross-domain test set consisting of two subsets: (1) a GAN-based set that includes synthetic images from ProGAN, StyleGAN, BigGAN, CycleGAN, StarGAN, GauGAN, and WFIR, all provided by CNNSpot; (2) a diffusion-based set adopted from~\cite{intro_04, exper_RPTC}, which comprises samples from ADM, Glide, Midjourney, Stable Diffusion v1.4 and v1.5, VQDM, Wukong, DALLE$\cdot$2, and SD-XL.

\begin{table*}[t]
\centering
\small

\begin{subtable}[t]{.17\textwidth}
\centering

\begin{tabular}{@{\hskip 1pt}lcc@{\hskip 1pt}}
\toprule
Method & Acc & AP \\
\midrule
Original & 83.1 & 89.1 \\
Reconst. & 72.6 & 78.4 \\
Concat & 88.6 & 96.3 \\
\rowcolor{gray!20}\textbf{Residual} & \textbf{90.7} & \textbf{97.5} \\
\bottomrule
\end{tabular}
\caption{Aggregation method}
\end{subtable}
\hspace*{1.2em}
\begin{subtable}[t]{.17\textwidth}
\centering

\begin{tabular}{@{\hskip 1pt}lcc@{\hskip 1pt}}
\toprule
Number & Acc & AP \\
\midrule
1 & 81.3 & 87.3 \\
2 & 86.7 & 91.7 \\
4 & 89.4 & 96.3 \\
\rowcolor{gray!20}\textbf{6} & \textbf{90.7} & \textbf{97.5} \\
\bottomrule
\end{tabular}
\caption{Granularity levels}
\end{subtable}
\hspace*{0.5em}
\begin{subtable}[t]{.17\textwidth}
\centering

\begin{tabular}{@{\hskip 1pt}lcc@{\hskip 1pt}}
\toprule
Number & Acc & AP \\
\midrule
1 & 85.3 & 89.0 \\
2 & 87.4 & 93.1 \\
\rowcolor{gray!20}\textbf{3} & \textbf{90.7} & \textbf{97.5} \\
4 & 91.0 & 96.9 \\
\bottomrule
\end{tabular}
\caption{Manifold branches}
\end{subtable}
\hspace*{0.6em}
\begin{subtable}[t]{.17\textwidth}
\centering

\begin{tabular}{@{\hskip 1pt}lcc@{\hskip 1pt}}
\toprule
Extractor & Acc & AP \\
\midrule
Vanilla CLIP & 73.5 & 79.4 \\
ResNet50 & 81.6 & 86.2 \\
finetune CLIP & 86.3 & 92.2 \\
\rowcolor{gray!20}\textbf{Ours} & \textbf{90.7} & \textbf{97.5} \\
\bottomrule
\end{tabular}
\caption{GAE feature type}
\end{subtable}
\hspace*{2.8em}
\begin{subtable}[t]{.17\textwidth}
\centering

\begin{tabular}{@{\hskip 1pt}lcc@{\hskip 1pt}}
\toprule
Domain & Acc & AP \\
\midrule
RGB & 87.1 & 94.2 \\
SRM & 88.2 & 94.6 \\
DFT & 87.8 & 93.2 \\
\rowcolor{gray!20}\textbf{All} & \textbf{90.7} & \textbf{97.5} \\
\bottomrule
\end{tabular}
\caption{MLRE domain input}
\end{subtable}

\caption{Ablation study on key design components. We report AP and Acc (\%).}
\label{tab:4}
\end{table*}

\begin{table}[t]

  \centering
  \resizebox{\columnwidth}{!}{
  \begin{tabular}{>{\raggedright\arraybackslash}m{2cm}>{\centering\arraybackslash}m{1.4cm}>{\centering\arraybackslash}m{1.4cm}>{\centering\arraybackslash}m{1.4cm}>{\centering\arraybackslash}m{1.4cm}}
  \toprule
 Methods & Blurring & Cropping & JPEG & Noise \\
  \midrule
CNNSpot   &  63.3/70.1 &  61.3/67.7 &   64.4/69.1 &  62.6/70.4 \\
FreDect   &  60.5/62.9 &  59.2/62.7 &   59.1/63.6 &  58.7/62.3 \\
GramNet  &  63.6/64.5 &  60.7/63.3 &   61.0/64.1 &  62.1/64.2 \\
LGrad    &  67.7/75.1 &  63.3/74.0 &   67.6/74.7 &  64.8/75.7 \\
UniFD     &  67.4/78.4 &  66.9/77.2 &   70.2/80.0 &  66.7/76.1 \\
DIRE     &  65.6/68.7 &  64.3/69.1 &   66.1/69.6 &  64.6/67.0 \\
AEROBLADE  &  47.0/60.2 &  48.0/62.7 &   49.7/63.4 &  46.6/62.2 \\
NPR      &  77.8/80.8 &  76.7/79.9 &   78.5/81.6 &  76.2/79.2 \\

Fatformer &  81.6/86.0 &  81.2/86.6 &   82.6/85.6 &  79.2/84.0 \\
DRCT      &  79.8/87.4 &  80.3/85.6 &   81.8/87.1 &  78.5/86.0 \\

D$^3$  & 80.6/85.6&80.8/86.4&81.5/86.6&80.4/85.1\\
CO-SPY      &  \underline{83.7}/\underline{88.3} & \underline{83.3}/\underline{87.7} &   \underline{84.2}/\underline{87.5} &  \underline{82.1}/\underline{86.5}  \\
\rowcolor{gray!20}Ours  &  \textbf{86.3}/\textbf{91.1} &  \textbf{85.6}/\textbf{89.9} &   \textbf{87.2}/\textbf{90.3} &  \textbf{85.0}/\textbf{89.2} \\
  \bottomrule
  \end{tabular}}
  \caption{Robustness evaluation. The average Acc/AP scores (\%) across all test data are reported.}
  \label{tab:3}
\end{table}

\noindent \textbf{Evaluation Metric.}
Following standard practice~\cite{method_14, method_12}, we use accuracy (Acc) and average precision (AP) as metrics, and report their means (mAcc, mAP) for comprehensive assessment.

\noindent \textbf{Implementation Details.}
We follow standardized protocols from prior works~\cite{method_14, method_12}.
Each expert employs a pre-trained VGG-16~\cite{vgg} as the HGE for multi-scale feature extraction, followed by lightweight CNN network~\cite{tan2024rethinking} for final prediction.
Experts are divided into six subgroups based on feature scale, with each subgroup comprising three heterogeneous experts initialized from autoencoders pre-trained on Kandinsky 2.1~\cite{kd}, Stable Diffusion 1.1, and 2.1~\cite{intro_311}.

The loss-balancing coefficients \(\alpha\) and \(\beta\) are empirically set to 0.5 and \(1 \times 10^{-2}\), respectively.
Training is performed for 10 epochs with a batch size of 64, using SGD with a learning rate of \(1 \times 10^{-4}\) and momentum of 0.9.
All experiments are run on an A800 GPU. Additional implementation details are available in the supplementary material.

\subsection{Comparison Results}
In this section, we compare TrueMoE with several state-of-the-art synthetic image detection methods, including CNNSpot~\cite{method_14}, FreDect~\cite{method_26}, GramNet~\cite{method_16}, LGrad~\cite{method_13}, UniFD~\cite{method_12}, DIRE~\cite{method_11}, AEROBLADE~\cite{aeroblade}, NPR~\cite{tan2024rethinking}, FatFormer~\cite{fatformer}, DRCT~\cite{drct}, D$^3$~\cite{D3}, and CO-SPY~\cite{co-spy}.
To ensure a fair comparison, we re-trained all baseline models exclusively on ProGAN-generated images using a standardized training protocol consistent with our method.
For DRCT, which requires specialized training pipelines, we use their official weights.
Since AEROBLADE does not involve training, we normalize its distance-based scores and apply a fixed threshold (0.5) to compute accuracy.
As shown in Table~\ref{tab:1} and Table~\ref{tab:2}, TrueMoE achieves the best overall performance, surpassing the strongest baselines by +2.98\% in mAcc and +2.04\% in mAP.
These results highlight the effectiveness of our dual-routing and multi-expert architecture, which adaptively selects experts based on manifold representations and granularity-aware features.

Unlike prior methods that build a single generalizable feature space, TrueMoE decomposes detection into analysis and collaborative expert decision-making, leveraging expert diversity to enhance generalization.
In contrast, unified detection spaces often fail to adapt to distribution shifts.
For example, CO-SPY combines high-level semantics and low-level artifacts in a single framework, performs well on GAN images but degrades significantly on diffusion-generated content due to ProGAN-specific training.
Similarly, although AEROBLADE uses autoencoder-based features, it struggles with generators beyond latent diffusion models.
In contrast, TrueMoE leverages the MoE paradigm to integrate diverse experts and aggregate multi-level features across hybrid manifolds.
As a result, it generalizes well across diverse generative models.
We also provide visualizations in the supplementary material to further illustrate the design motivation and effectiveness.

\subsection{Robustness Evaluation}
To evaluate TrueMoE’s robustness, we follow the protocol from FreDect and FatFormer, applying four distortions each with a 50\% probability: (1) Gaussian blur with kernel sizes randomly chosen from (3, 5, 7, 9); (2) random cropping along horizontal and vertical axes; (3) JPEG compression with variable quality factors; and (4) additive Gaussian noise with variance sampled from the range [5, 20].
As shown in Table~\ref{tab:3}, TrueMoE consistently achieves the highest accuracy and AP under all perturbations, outperforming strong baselines like CO-SPY and DRCT.
For instance, under JPEG compression, CO-SPY achieves 84.2\% / 87.5\%, falling short by 3.0\% / 2.8\%\ compared to TrueMoE.
This robustness stems from the large and diverse detection space we construct. The Granularity-Aware Artifact Extractor and Multi-domain Latent Representation Extractor suppress semantic content and extract robust, forgery-relevant features, enabling accurate expert selection under distribution shifts.

\subsection{Ablation Study}

\noindent \textbf{Effect of Feature Aggregation Strategies.}
We evaluate several strategies for constructing the Granularity-Aware Discrimination Feature.
Table~\ref{tab:3}(a) shows that residual features yield the best performance,  as they effectively reveal artifacts introduced by generative models.  
Accordingly, we adopt residual features as input to the expert discriminators.

\noindent \textbf{Effect of Expert Number in DEA.}
We ablate the number of experts in DEA by varying the two structural axes: granularity and manifold diversity (Table~\ref{tab:4}(b)(c)).  
 As shown in the results, increasing the expert count consistently improves detection performance.  
This confirms that a larger and more diverse ensemble of specialized experts enables finer characterization of generative artifacts.
While adding experts yields steady gains, we observe diminishing returns and increased cost beyond six granularity levels and three manifold branches, which are used as the default setting.

\noindent \textbf{Effect of Granularity-Aware Pretraining.}
To validate the effectiveness of our pretraining strategy for the Granularity-Aware Artifact Extractor, we replace it with several alternative feature extractors.
As shown in Table~\ref{tab:4}(d), the vanilla CLIP model fails to separate forgery-specific signals from high-level semantics, leading to a substantial drop in performance (i.e., ↓18.1\% AP). Even with fine-tuning, CLIP only achieves moderate improvement.
This degradation occurs due to the difficulty of disentangling high-level semantics from forgery-specific signals using conventional supervised learning.
In contrast, our contrastive pretraining jointly optimizes semantic and granularity objectives, resulting in more discriminative and artifact-sensitive embeddings.

\noindent \textbf{Effect of Multi-Domain Representation.}
The RGB, SRM, and DFT domains respectively capture spatial structures, local noise patterns, and frequency cues. Their integration enables comprehensive manifold modeling. As evidenced by Table~\ref{tab:4}(e), removing any single domain leads to a consistent performance drop, confirming their complementary roles in facilitating robust cross-domain representation.

Ablation study reveals that the number of experts and the quality of pretrained routing features from GAE are the two most critical factors for performance improvement.
\section{Conclusion}

In this paper, we presented TrueMoE, a novel Mixture-of-Experts framework tailored for generalizable synthetic image detection. By organizing experts along manifold and granularity axes and introducing a dual-routing mechanism, our model adaptively selects appropriate expert combinations for each input. Extensive experiments demonstrate the effectiveness of our design, showing superior performance and robustness across diverse generative models, especially under unseen settings and challenging degradations. However, despite its strong generalization ability, TrueMoE still requires further improvements, such as reducing training costs, enhancing scalability, and better handling extremely subtle forgeries.

\bibliography{TrueMoE}

\begin{thebibliography}{55}
\providecommand{\natexlab}[1]{#1}

\bibitem[{Artetxe et~al.(2021)Artetxe, Bhosale, Goyal, Mihaylov, Ott, Shleifer, Lin, Du, Iyer, Pasunuru et~al.}]{artetxe2021efficient}
Artetxe, M.; Bhosale, S.; Goyal, N.; Mihaylov, T.; Ott, M.; Shleifer, S.; Lin, X.~V.; Du, J.; Iyer, S.; Pasunuru, R.; et~al. 2021.
\newblock {Efficient large scale language modeling with mixtures of experts}.
\newblock \emph{arXiv preprint arXiv:2112.10684}.

\bibitem[{Brock, Donahue, and Simonyan(2019)}]{intro_302}
Brock, A.; Donahue, J.; and Simonyan, K. 2019.
\newblock Large Scale {GAN} Training for High Fidelity Natural Image Synthesis.
\newblock In \emph{International Conference on Learning Representations}.

\bibitem[{Chang et~al.(2020)Chang, Lan, Cheng, and Wei}]{uncertain}
Chang, J.; Lan, Z.; Cheng, C.; and Wei, Y. 2020.
\newblock Data uncertainty learning in face recognition.
\newblock In \emph{Proceedings of the IEEE/CVF Conference on Computer Vision and Pattern Recognition}, 5710--5719.

\bibitem[{Chen et~al.(2024)Chen, Zeng, Yang, and Yang}]{drct}
Chen, B.; Zeng, J.; Yang, J.; and Yang, R. 2024.
\newblock Drct: Diffusion reconstruction contrastive training towards universal detection of diffusion generated images.
\newblock In \emph{Forty-first International Conference on Machine Learning}.

\bibitem[{Cheng et~al.(2025)Cheng, Lyu, Wang, Zhang, and Sehwag}]{co-spy}
Cheng, S.; Lyu, L.; Wang, Z.; Zhang, X.; and Sehwag, V. 2025.
\newblock CO-SPY: Combining Semantic and Pixel Features to Detect Synthetic Images by AI.
\newblock In \emph{Proceedings of the Computer Vision and Pattern Recognition Conference}, 13455--13465.

\bibitem[{Choi et~al.(2018)Choi, Choi, Kim, Ha, Kim, and Choo}]{intro_304}
Choi, Y.; Choi, M.; Kim, M.; Ha, J.-W.; Kim, S.; and Choo, J. 2018.
\newblock StarGAN: Unified Generative Adversarial Networks for Multi-Domain Image-to-Image Translation.
\newblock In \emph{Proceedings of the IEEE Conference on Computer Vision and Pattern Recognition}.

\bibitem[{Chowdhury et~al.(2023)Chowdhury, Zhang, Wang, Liu, and Chen}]{pmoe}
Chowdhury, M. N.~R.; Zhang, S.; Wang, M.; Liu, S.; and Chen, P.-Y. 2023.
\newblock {Patch-level routing in mixture-of-experts is provably sample-efficient for convolutional neural networks}.
\newblock In \emph{International Conference on Machine Learning}, 6074--6114. PMLR.

\bibitem[{Costa-juss{\`a} et~al.(2022)Costa-juss{\`a}, Cross, {\c{C}}elebi, Elbayad, Heafield, Heffernan, Kalbassi, Lam, Licht, Maillard et~al.}]{costa2022no}
Costa-juss{\`a}, M.~R.; Cross, J.; {\c{C}}elebi, O.; Elbayad, M.; Heafield, K.; Heffernan, K.; Kalbassi, E.; Lam, J.; Licht, D.; Maillard, J.; et~al. 2022.
\newblock {No language left behind: Scaling human-centered machine translation}.
\newblock \emph{arXiv preprint arXiv:2207.04672}.

\bibitem[{Dai et~al.(2024)Dai, Deng, Zhao, Xu, Gao, Chen, Li, Zeng, Yu, Wu et~al.}]{dai2024deepseekmoe}
Dai, D.; Deng, C.; Zhao, C.; Xu, R.; Gao, H.; Chen, D.; Li, J.; Zeng, W.; Yu, X.; Wu, Y.; et~al. 2024.
\newblock {Deepseekmoe: Towards ultimate expert specialization in mixture-of-experts language models}.
\newblock \emph{arXiv preprint arXiv:2401.06066}.

\bibitem[{Dhariwal and Nichol(2021)}]{intro_307}
Dhariwal, P.; and Nichol, A. 2021.
\newblock Diffusion Models Beat GANs on Image Synthesis.
\newblock In \emph{Advances in Neural Information Processing Systems}, 8780--8794.

\bibitem[{Du et~al.(2022)Du, Huang, Dai, Tong, Lepikhin, Xu, Krikun, Zhou, Yu, Firat et~al.}]{du2022glam}
Du, N.; Huang, Y.; Dai, A.~M.; Tong, S.; Lepikhin, D.; Xu, Y.; Krikun, M.; Zhou, Y.; Yu, A.~W.; Firat, O.; et~al. 2022.
\newblock {Glam: Efficient scaling of language models with mixture-of-experts}.
\newblock In \emph{International Conference on Machine Learning}, 5547--5569. PMLR.

\bibitem[{Frank et~al.(2020)Frank, Eisenhofer, Sch{\"o}nherr, Fischer, Kolossa, and Holz}]{method_26}
Frank, J.; Eisenhofer, T.; Sch{\"o}nherr, L.; Fischer, A.; Kolossa, D.; and Holz, T. 2020.
\newblock Leveraging Frequency Analysis for Deep Fake Image Recognition.
\newblock In \emph{Proceedings of the 37th International Conference on Machine Learning}, 3247--3258.

\bibitem[{Fridrich and Kodovsky(2012)}]{srm}
Fridrich, J.; and Kodovsky, J. 2012.
\newblock Rich models for steganalysis of digital images.
\newblock \emph{IEEE Transactions on information Forensics and Security}, 7(3): 868--882.

\bibitem[{Gu et~al.(2022)Gu, Chen, Bao, Wen, Zhang, Chen, Yuan, and Guo}]{intro_312}
Gu, S.; Chen, D.; Bao, J.; Wen, F.; Zhang, B.; Chen, D.; Yuan, L.; and Guo, B. 2022.
\newblock Vector Quantized Diffusion Model for Text-to-Image Synthesis.
\newblock In \emph{Proceedings of the IEEE/CVF Conference on Computer Vision and Pattern Recognition}, 10696--10706.

\bibitem[{Guo et~al.(2023)Guo, Liu, Ren, Grosz, Masi, and Liu}]{guo2023hierarchical}
Guo, X.; Liu, X.; Ren, Z.; Grosz, S.; Masi, I.; and Liu, X. 2023.
\newblock Hierarchical fine-grained image forgery detection and localization.
\newblock In \emph{Proceedings of the IEEE/CVF Conference on Computer Vision and Pattern Recognition}, 3155--3165.

\bibitem[{Hu et~al.(2021)Hu, Shen, Wallis, Allen-Zhu, Li, Wang, and Chen}]{lora}
Hu, E.~J.; Shen, Y.; Wallis, P.; Allen-Zhu, Z.; Li, Y.; Wang, S.; and Chen, W. 2021.
\newblock LoRA: Low-Rank Adaptation of Large Language Models.
\newblock arXiv:2106.09685.

\bibitem[{Jacobs et~al.(1991)Jacobs, Jordan, Nowlan, and Hinton}]{moe}
Jacobs, R.~A.; Jordan, M.~I.; Nowlan, S.~J.; and Hinton, G.~E. 1991.
\newblock Adaptive mixtures of local experts.
\newblock \emph{Neural computation}, 3(1): 79--87.

\bibitem[{Jiang et~al.(2024)Jiang, Sablayrolles, Roux, Mensch, Savary, Bamford, Chaplot, Casas, Hanna, Bressand et~al.}]{jiang2024mixtral}
Jiang, A.~Q.; Sablayrolles, A.; Roux, A.; Mensch, A.; Savary, B.; Bamford, C.; Chaplot, D.~S.; Casas, D. d.~l.; Hanna, E.~B.; Bressand, F.; et~al. 2024.
\newblock {Mixtral of experts}.
\newblock \emph{arXiv preprint arXiv:2401.04088}.

\bibitem[{Karras et~al.(2018)Karras, Aila, Laine, and Lehtinen}]{intro_301}
Karras, T.; Aila, T.; Laine, S.; and Lehtinen, J. 2018.
\newblock Progressive Growing of GANs for Improved Quality, Stability, and Variation.
\newblock In \emph{Proceedings of International Conference on Learning Representations}.

\bibitem[{Kietzmann et~al.(2020)Kietzmann, Lee, McCarthy, and Kietzmann}]{kietzmann2020deepfakes}
Kietzmann, J.; Lee, L.~W.; McCarthy, I.~P.; and Kietzmann, T.~C. 2020.
\newblock Deepfakes: Trick or treat?
\newblock \emph{Business Horizons}, 63(2): 135--146.

\bibitem[{Lepikhin et~al.(2020)Lepikhin, Lee, Xu, Chen, Firat, Huang, Krikun, Shazeer, and Chen}]{lepikhin2020gshard}
Lepikhin, D.; Lee, H.; Xu, Y.; Chen, D.; Firat, O.; Huang, Y.; Krikun, M.; Shazeer, N.; and Chen, Z. 2020.
\newblock Gshard: Scaling giant models with conditional computation and automatic sharding.
\newblock \emph{arXiv preprint arXiv:2006.16668}.

\bibitem[{Li et~al.(2022)Li, Li, Xiong, and Hoi}]{blip}
Li, J.; Li, D.; Xiong, C.; and Hoi, S. 2022.
\newblock Blip: Bootstrapping language-image pre-training for unified vision-language understanding and generation.
\newblock In \emph{International conference on machine learning}, 12888--12900. PMLR.

\bibitem[{Li et~al.(2024{\natexlab{a}})Li, Cai, Hao, Jiang, Hu, and Feng}]{li2024improving}
Li, O.; Cai, J.; Hao, Y.; Jiang, X.; Hu, Y.; and Feng, F. 2024{\natexlab{a}}.
\newblock Improving Synthetic Image Detection Towards Generalization: An Image Transformation Perspective.
\newblock \emph{arXiv preprint arXiv:2408.06741}.

\bibitem[{Li et~al.(2024{\natexlab{b}})Li, Ma, Guo, Xu, Li, and Zhang}]{li2024unionformer}
Li, S.; Ma, W.; Guo, J.; Xu, S.; Li, B.; and Zhang, X. 2024{\natexlab{b}}.
\newblock Unionformer: Unified-learning transformer with multi-view representation for image manipulation detection and localization.
\newblock In \emph{Proceedings of the IEEE/CVF Conference on Computer Vision and Pattern Recognition}, 12523--12533.

\bibitem[{Liu et~al.(2024{\natexlab{a}})Liu, Feng, Wang, Wang, Liu, Zhao, Dengr, Ruan, Dai, Guo et~al.}]{liu2024deepseek}
Liu, A.; Feng, B.; Wang, B.; Wang, B.; Liu, B.; Zhao, C.; Dengr, C.; Ruan, C.; Dai, D.; Guo, D.; et~al. 2024{\natexlab{a}}.
\newblock Deepseek-v2: A strong, economical, and efficient mixture-of-experts language model.
\newblock \emph{arXiv preprint arXiv:2405.04434}.

\bibitem[{Liu et~al.(2024{\natexlab{b}})Liu, Tan, Tan, Wei, Wang, and Zhao}]{fatformer}
Liu, H.; Tan, Z.; Tan, C.; Wei, Y.; Wang, J.; and Zhao, Y. 2024{\natexlab{b}}.
\newblock Forgery-aware Adaptive Transformer for Generalizable Synthetic Image Detection.
\newblock In \emph{Proceedings of the IEEE/CVF Conference on Computer Vision and Pattern Recognition}.

\bibitem[{Liu, Qi, and Torr(2020)}]{method_16}
Liu, Z.; Qi, X.; and Torr, P.~H. 2020.
\newblock Global Texture Enhancement for Fake Face Detection in the Wild.
\newblock In \emph{Proceedings of the IEEE/CVF Conference on Computer Vision and Pattern Recognition}.

\bibitem[{MacQueen(1967)}]{kmean}
MacQueen, J. 1967.
\newblock Some methods for classification and analysis of multivariate observations.
\newblock In \emph{Proceedings of the Fifth Berkeley Symposium on Mathematical Statistics and Probability, Volume 1: Statistics}, volume~5, 281--298. University of California press.

\bibitem[{Midjourney(2023)}]{intro_309}
Midjourney. 2023.
\newblock https://www.midjourney.com/home/.

\bibitem[{Nichol et~al.(2022)Nichol, Dhariwal, Ramesh, Shyam, Mishkin, Mcgrew, Sutskever, and Chen}]{intro_308}
Nichol, A.~Q.; Dhariwal, P.; Ramesh, A.; Shyam, P.; Mishkin, P.; Mcgrew, B.; Sutskever, I.; and Chen, M. 2022.
\newblock {GLIDE}: Towards Photorealistic Image Generation and Editing with Text-Guided Diffusion Models.
\newblock In \emph{Proceedings of the 39th International Conference on Machine Learning}, 16784--16804.

\bibitem[{Ojha, Li, and Lee(2023)}]{method_12}
Ojha, U.; Li, Y.; and Lee, Y.~J. 2023.
\newblock Towards Universal Fake Image Detectors That Generalize Across Generative Models.
\newblock In \emph{Proceedings of the IEEE/CVF Conference on Computer Vision and Pattern Recognition}, 24480--24489.

\bibitem[{Qian et~al.(2020)Qian, Yin, Sheng, Chen, and Shao}]{method_114}
Qian, Y.; Yin, G.; Sheng, L.; Chen, Z.; and Shao, J. 2020.
\newblock Thinking in Frequency: Face Forgery Detection by Mining Frequency-Aware Clues.
\newblock In \emph{European Conference on Computer Vision}, 86--103.

\bibitem[{Radford et~al.(2021)Radford, Kim, Hallacy, Ramesh, Goh, Agarwal, Sastry, Askell, Mishkin, Clark et~al.}]{radford2021learning}
Radford, A.; Kim, J.~W.; Hallacy, C.; Ramesh, A.; Goh, G.; Agarwal, S.; Sastry, G.; Askell, A.; Mishkin, P.; Clark, J.; et~al. 2021.
\newblock Learning transferable visual models from natural language supervision.
\newblock In \emph{International conference on machine learning}, 8748--8763. PMLR.

\bibitem[{Razzhigaev et~al.(2023)Razzhigaev, Shakhmatov, Maltseva, Arkhipkin, Pavlov, Ryabov, Kuts, Panchenko, Kuznetsov, and Dimitrov}]{kd}
Razzhigaev, A.; Shakhmatov, A.; Maltseva, A.; Arkhipkin, V.; Pavlov, I.; Ryabov, I.; Kuts, A.; Panchenko, A.; Kuznetsov, A.; and Dimitrov, D. 2023.
\newblock Kandinsky: an improved text-to-image synthesis with image prior and latent diffusion.
\newblock \emph{arXiv preprint arXiv:2310.03502}.

\bibitem[{Ricker et~al.(2024)Ricker, Lukovnikov, Fischer, and Asja}]{aeroblade}
Ricker, J.; Lukovnikov, D.; Fischer; and Asja. 2024.
\newblock Aeroblade: Training-free detection of latent diffusion images using autoencoder reconstruction error.
\newblock In \emph{Proceedings of the IEEE/CVF Conference on Computer Vision and Pattern Recognition}, 9130--9140.

\bibitem[{Riquelme et~al.(2021)Riquelme, Puigcerver, Mustafa, Neumann, Jenatton, Susano~Pinto, Keysers, and Houlsby}]{2.3.2}
Riquelme, C.; Puigcerver, J.; Mustafa, B.; Neumann, M.; Jenatton, R.; Susano~Pinto, A.; Keysers, D.; and Houlsby, N. 2021.
\newblock Scaling vision with sparse mixture of experts.
\newblock \emph{Advances in Neural Information Processing Systems}, 34: 8583--8595.

\bibitem[{Rombach et~al.(2022)Rombach, Blattmann, Lorenz, Esser, and Ommer}]{intro_311}
Rombach, R.; Blattmann, A.; Lorenz, D.; Esser, P.; and Ommer, B. 2022.
\newblock High-Resolution Image Synthesis With Latent Diffusion Models.
\newblock In \emph{Proceedings of the IEEE/CVF Conference on Computer Vision and Pattern Recognition}, 10684--10695.

\bibitem[{Shazeer et~al.(2017)Shazeer, Mirhoseini, Maziarz, Davis, Le, Hinton, and Dean}]{shazeer2017outrageously}
Shazeer, N.; Mirhoseini, A.; Maziarz, K.; Davis, A.; Le, Q.; Hinton, G.; and Dean, J. 2017.
\newblock {Outrageously large neural networks: The sparsely-gated mixture-of-experts layer}.
\newblock \emph{arXiv preprint arXiv:1701.06538}.

\bibitem[{Simonyan and Zisserman(2014)}]{vgg}
Simonyan, K.; and Zisserman, A. 2014.
\newblock Very deep convolutional networks for large-scale image recognition.
\newblock \emph{arXiv preprint arXiv:1409.1556}.

\bibitem[{Singh et~al.(2023)Singh, Ruwase, Awan, Rajbhandari, He, and Bhatele}]{2.3.3}
Singh, S.; Ruwase, O.; Awan, A.~A.; Rajbhandari, S.; He, Y.; and Bhatele, A. 2023.
\newblock A hybrid tensor-expert-data parallelism approach to optimize mixture-of-experts training.
\newblock In \emph{Proceedings of the 37th International Conference on Supercomputing}, 203--214.

\bibitem[{Tan et~al.(2024)Tan, Zhao, Wei, Gu, Liu, and Wei}]{tan2024rethinking}
Tan, C.; Zhao, Y.; Wei, S.; Gu, G.; Liu, P.; and Wei, Y. 2024.
\newblock Rethinking the up-sampling operations in cnn-based generative network for generalizable deepfake detection.
\newblock In \emph{Proceedings of the IEEE/CVF Conference on Computer Vision and Pattern Recognition}, 28130--28139.

\bibitem[{Tan et~al.(2023)Tan, Zhao, Wei, Gu, and Wei}]{method_13}
Tan, C.; Zhao, Y.; Wei, S.; Gu, G.; and Wei, Y. 2023.
\newblock Learning on Gradients: Generalized Artifacts Representation for GAN-Generated Images Detection.
\newblock In \emph{Proceedings of the IEEE/CVF Conference on Computer Vision and Pattern Recognition}, 12105--12114.

\bibitem[{Vincent(2023)}]{vincent2023pope}
Vincent, J. 2023.
\newblock The swagged-out pope is an AI fake — and an early glimpse of a new reality.
\newblock \textit{The Verge}.

\bibitem[{Wang et~al.(2022)Wang, Wu, Ouyang, Han, Chen, Jiang, and Li}]{wang2022m2tr}
Wang, J.; Wu, Z.; Ouyang, W.; Han, X.; Chen, J.; Jiang, Y.-G.; and Li, S.-N. 2022.
\newblock M2tr: Multi-modal multi-scale transformers for deepfake detection.
\newblock In \emph{Proceedings of the 2022 international conference on multimedia retrieval}, 615--623.

\bibitem[{Wang et~al.(2020)Wang, Wang, Zhang, Owens, and Efros}]{method_14}
Wang, S.-Y.; Wang, O.; Zhang, R.; Owens, A.; and Efros, A.~A. 2020.
\newblock CNN-Generated Images Are Surprisingly Easy to Spot... for Now.
\newblock In \emph{Proceedings of the IEEE/CVF Conference on Computer Vision and Pattern Recognition}.

\bibitem[{Wang et~al.(2023)Wang, Bao, Zhou, Wang, Hu, Chen, and Li}]{method_11}
Wang, Z.; Bao, J.; Zhou, W.; Wang, W.; Hu, H.; Chen, H.; and Li, H. 2023.
\newblock DIRE for Diffusion-Generated Image Detection.
\newblock In \emph{Proceedings of the IEEE/CVF International Conference on Computer Vision}, 22445--22455.

\bibitem[{Wang et~al.(2024)Wang, Sehwag, Chen, Lyu, Metaxas, and Ma}]{wang2024trace}
Wang, Z.; Sehwag, V.; Chen, C.; Lyu, L.; Metaxas, D.~N.; and Ma, S. 2024.
\newblock How to trace latent generative model generated images without artificial watermark?
\newblock \emph{arXiv preprint arXiv:2405.13360}.

\bibitem[{Xu, Fan, and Kankanhalli(2023)}]{xu2023combating}
Xu, D.; Fan, S.; and Kankanhalli, M. 2023.
\newblock Combating misinformation in the era of generative AI models.
\newblock In \emph{Proceedings of the 31st ACM International Conference on Multimedia}, 9291--9298.

\bibitem[{Yang et~al.(2025)Yang, Qian, Zhu, Russakovsky, and Wu}]{D3}
Yang, Y.; Qian, Z.; Zhu, Y.; Russakovsky, O.; and Wu, Y. 2025.
\newblock D\^{} 3: Scaling Up Deepfake Detection by Learning from Discrepancy.
\newblock In \emph{Proceedings of the Computer Vision and Pattern Recognition Conference}, 23850--23859.

\bibitem[{Zhang et~al.(2018)Zhang, Isola, Efros, Shechtman, and Wang}]{lpips}
Zhang, R.; Isola, P.; Efros, A.~A.; Shechtman, E.; and Wang, O. 2018.
\newblock The unreasonable effectiveness of deep features as a perceptual metric.
\newblock In \emph{Proceedings of the IEEE conference on computer vision and pattern recognition}, 586--595.

\bibitem[{{Zhong} et~al.(2023){Zhong}, {Xu}, {Li}, {Qian}, and {Zhang}}]{exper_RPTC}
{Zhong}, N.; {Xu}, Y.; {Li}, S.; {Qian}, Z.; and {Zhang}, X. 2023.
\newblock {PatchCraft: Exploring Texture Patch for Efficient AI-generated Image Detection}.
\newblock \emph{arXiv e-prints}, arXiv:2311.12397.

\bibitem[{Zhou and Zhou(2021)}]{zhou2021ensemble}
Zhou, Z.-H.; and Zhou, Z.-H. 2021.
\newblock \emph{Ensemble learning}.
\newblock Springer.

\bibitem[{Zhu et~al.(2017)Zhu, Park, Isola, and Efros}]{intro_303}
Zhu, J.-Y.; Park, T.; Isola, P.; and Efros, A.~A. 2017.
\newblock Unpaired Image-To-Image Translation Using Cycle-Consistent Adversarial Networks.
\newblock In \emph{Proceedings of the IEEE International Conference on Computer Vision}.

\bibitem[{Zhu et~al.(2023)Zhu, Chen, YAN, Huang, Lin, Li, Tu, Hu, Hu, and Wang}]{intro_04}
Zhu, M.; Chen, H.; YAN, Q.; Huang, X.; Lin, G.; Li, W.; Tu, Z.; Hu, H.; Hu, J.; and Wang, Y. 2023.
\newblock GenImage: A Million-Scale Benchmark for Detecting AI-Generated Image.
\newblock In \emph{Advances in Neural Information Processing Systems}, volume~36, 77771--77782.

\bibitem[{Zounemat-Kermani et~al.(2021)Zounemat-Kermani, Batelaan, Fadaee, and Hinkelmann}]{zounemat2021ensemble}
Zounemat-Kermani, M.; Batelaan, O.; Fadaee, M.; and Hinkelmann, R. 2021.
\newblock Ensemble machine learning paradigms in hydrology: A review.
\newblock \emph{Journal of Hydrology}, 598: 126266.

\end{thebibliography}

\end{document}